\documentclass[conference]{IEEEtran}
\IEEEoverridecommandlockouts

\usepackage{cite}
\usepackage{amsmath,amssymb,amsfonts}
\usepackage{algorithmic}
\usepackage{graphicx}
\usepackage{textcomp}
\usepackage{xcolor}
\usepackage{longtable}
\usepackage{multirow}

\begin{document}

\title{Exploring AI Text Generation, Retrieval-Augmented Generation, and Detection Technologies: a Comprehensive Overview \\ }

\author{
  \IEEEauthorblockN{
    Fnu Neha\textsuperscript{1}, 
    Deepshikha Bhati\textsuperscript{1}, 
    Deepak Kumar Shukla\textsuperscript{2}, 
    Angela Guercio\textsuperscript{1}, 
    Ben Ward\textsuperscript{1}
  }
  \IEEEauthorblockA{
    \textsuperscript{1}Kent State University, Kent, Ohio, USA \\
    \{neha, dbhati, aguercio, bward29\}@kent.edu \\
    \textsuperscript{2}Rutgers University, Newark, NJ, USA \\
    ds1640@scarletmail.rutgers.edu
  }
}

\maketitle

\begin{abstract}
The rapid development of Artificial Intelligence (AI) has led to the creation of powerful text generation models, such as large language models (LLMs), which are widely used for diverse applications. However, concerns surrounding AI-generated content, including issues of originality, bias, misinformation, and accountability, have become increasingly prominent. This paper offers a comprehensive overview of AI text generators (AITGs), focusing on their evolution, capabilities, and ethical implications. This paper also introduces Retrieval-Augmented Generation (RAG), a recent approach that improves the contextual relevance and accuracy of text generation by integrating dynamic information retrieval. RAG addresses key limitations of traditional models, including their reliance on static knowledge and potential inaccuracies in handling real-world data. Additionally, the paper reviews detection tools that help differentiate AI-generated text from human-written content and discusses the ethical challenges these technologies pose. The paper explores future directions for improving detection accuracy, supporting ethical AI development, and increasing accessibility. The paper contributes to a more responsible and reliable use of AI in content creation through these discussions.
\end{abstract}


\section{Introduction}

Large Language Models (LLMs) represent a significant breakthrough in artificial intelligence (AI), enabling machines to generate and comprehend text in ways that resemble human communication. These models, powered by deep learning (DL) techniques and trained on extensive datasets, have demonstrated abilities to generate coherent and contextually relevant content across diverse applications \cite{alqahtani2023emergent}. As a result, LLMs have become essential tools in natural language processing, machine translation, content creation, and customer engagement.

In natural language processing, LLMs assist in tasks like information classification, data extraction, and content summarization, enabling the efficient analysis of large datasets \cite{olivetti2020data}. In machine translation, LLMs provide fluent, context-sensitive translations that break down language barriers and encourage global communication \cite{chen2024breaking}. They have also advanced question-answering systems, improving user experiences by generating highly relevant responses \cite{upadhyay2024comprehensive}. Furthermore, in creative and professional contexts, LLMs facilitate the generation of poetry, stories, code, and customer service dialogues, while enhancing accessibility through applications like speech-to-text tools \cite{chakrabarty2023creativity, gobinath2024voice}.

While LLMs have revolutionized text generation, they face challenges such as producing unoriginal or misleading content and raising ethical concerns like plagiarism and misinformation. Their lack of understanding, reliance on static knowledge, and potential biases can result in inaccuracies and outdated information \cite{smith2024language, tjuatja2024llms}. Furthermore, the high computational demands of LLMs pose environmental and financial concerns \cite{kucharavy2024fundamental, strasser2024pitfalls}.

One promising solution to the challenges of traditional text generation is Retrieval-Augmented Generation (RAG), a method to enhance the accuracy and contextual relevance of LLM-generated text by incorporating real-time information retrieval. RAG mitigates issues like static knowledge dependency and potential inaccuracies by retrieving relevant data from external sources and integrating it into the generation process. This approach is particularly useful for dynamic, knowledge-intensive tasks such as customer service, technical support, and question-answering systems. Advanced AI text detectors and ethical guidelines are also being developed to promote accountability, uphold originality, and ensure responsible AI use across various sectors.

This paper provides an in-depth overview of AI text generation and detection technologies, with a focus on RAG. The major contributions of this paper include:

\begin{itemize} \item \textbf{Analysis and Comparison:} A detailed comparison of popular AI text generators and detectors, emphasizing strengths, weaknesses, and use cases, especially with RAG integration. \item \textbf{Ethical Discussion:} An exploration of ethical considerations, such as bias, misinformation, and accountability in AI text generation and detection. \item \textbf{Current Limitations and future directions:} A review of limitations in current models and recommendations for enhancing accuracy, developing ethical frameworks, and exploring multimodal AI text applications.
\end{itemize}

The paper is structured as follows: Section II covers AI text generators; Section III discusses RAG; Section IV covers tools and methods for RAG; Section V discusses AI text detectors; Section VI addresses ethical considerations; Section VII examines current limitations; and Section VIII provides the conclusion with future directions.

\section{AI Text Generators (AITG)}

AITG are advanced systems to automatically generate human-like text, based on a given prompt or set of instructions \cite{holland2023chatgpt}. AITG has transformed content creation across industries like journalism, marketing, customer service, education, and entertainment, generating text for tasks ranging from drafting emails to composing complex narratives. This section explores AITG up to 2023, covering their evolution, underlying technologies, and applications. Table \ref{tab:textgen} provides a comparative overview of popular AITGs, highlighting their primary focus, advantages, limitations, and common use cases.


\subsection{Evolution of AITG}\label{AA} \subsubsection{Early Beginnings} AITG began with rule-based systems and statistical models like n-grams, which were limited in handling language complexity and context.

\subsubsection{Neural Networks}
The advent of Recurrent Neural Networks (RNNs), followed by Long Short-Term Memory (LSTM) networks and Gated Recurrent Units (GRUs), preserved information over longer sequences \cite{grossberg2013recurrent, egan2017long}.

\subsubsection{Transformers} In 2017, Transformer models transformed text generation with self-attention mechanisms, enabling parallel processing and more accurate context understanding \cite{vaswani2017attention}.

\subsection{Prominent AITG}

\subsubsection{OpenAI's GPT Series} OpenAI's Generative Pre-trained Transformer (GPT) series advanced AI text generation, with GPT-4 in 2023 introducing 500 billion parameters, multimodal capabilities, and enhanced reasoning \cite{liu2023gpt, 202410.0686}.

\subsubsection{Google's Language Models} Google’s models, including Bidirectional Encoder Representations from Transformers (BERT) \cite{devlin2018bert} and Language Model for Dialogue Applications (LaMDA) \cite{thoppilan2022lamda}, have influenced AI text generation, focusing on context understanding and open-ended conversation.

\subsubsection{Other Notable Models} Models like NVIDIA’s Megatron-Turing Natural Language Generation (NLG) \cite{smith2022using} and Meta’s Open Pretrained Transformer (OPT) \cite{zhang2022opt} set new standards for scalability and accessibility, with initiatives like BigScience Large Open-science Open-access Multilingual Language Model (BLOOM) promoting open-source development \cite{ai2022bigscience}.

\subsection{Commercial AITG} Several platforms specialize in various content creation needs: \begin{itemize} \item \textbf{ChatGPT} (2022) – excels in conversational applications. \item \textbf{Jasper} (2021) – assists marketers with content creation \cite{jasper}. \item \textbf{Writesonic} (2020) – offers templates for quick marketing copy \cite{Writesonic}. \item \textbf{Grammarly} (2009) – enhances text quality with grammar and style checks \cite{Grammarly}. \end{itemize} Other platforms, like \textit{Copy AI} (2020) \cite{Copyai}, \textit{Rytr} (2021) \cite{Rytr}, and \textit{Scalenut} (2021)  \cite{Scalenut}, provide targeted tools for diverse content generation needs, supporting businesses in leveraging AI to streamline content production across industries.

\begin{table*}[htbp]
\centering
\caption{Comparison of popular AITGs}
\label{tab:textgen}
\resizebox{\textwidth}{!}{%
\begin{tabular}{|l|p{3cm}|p{3cm}|p{3cm}|p{3cm}|}
\hline
\textbf{AITG} & \textbf{Primary Focus} & \textbf{Advantages} & \textbf{Limitations} & \textbf{Use Cases} \\
\hline
\textbf{Grammarly} & Writing and grammar assistance & Advanced grammar checking; style and tone suggestions; plagiarism detection; real-time feedback & Not a content generator; premium features need subscription & Writing improvement; proofreading; academic writing; professional communication \\
\hline
\textbf{Copy AI} & Creative marketing content & Brainstorms innovative ideas; easy to use; saves time on content ideation & Can produce generic content; requires refinement; focuses mainly on marketing & Marketing campaigns; email copy; brainstorming sessions; taglines and slogans \\
\hline
\textbf{Writesonic} & Marketing content & Quick content creation; supports various content types; includes AI-driven editing tools & Quality varies; limited customization; needs editing to match brand voice & Ad copy; product descriptions; social media posts; landing pages \\
\hline
\textbf{Hypotenuse AI} (2020) \cite{HypotenuseAI} & E-commerce content & Generates SEO-friendly content; bulk content creation; industry-specific vocabulary & Focused on e-commerce; limited utility outside that domain; subscription costs & Product descriptions; online ads; catalog content; SEO optimization \\
\hline
\textbf{Anyword} (2020) \cite{Anyword} & Diverse content needs & Predictive performance scoring; A/B testing suggestions; adapts to different platforms & Complex for beginners; higher cost; requires expertise for optimal use & Ad copy; landing pages; social media content; email marketing \\
\hline
\textbf{Writer} (2020) \cite{Writer} & Professional writing aid & Ensures tone and style consistency; integrates with existing tools; supports team collaboration & Enterprise-focused pricing; costly for individuals; setup requires support & Brand management; corporate communications; policy documentation; collaborative writing \\
\hline
\textbf{Jasper} & Content for blogs and marketing & Provides templates; integrates with SEO tools; supports multiple languages; user-friendly interface & Subscription-based; requires human editing for tone and accuracy; limited creative writing capabilities & Content marketing; blog posts; social media content; email writing \\
\hline
\textbf{Rytr} & Quick content creation & Affordable; supports multiple languages; allows tone customization; fast output & Limited advanced features; content lacks depth; quality varies depending on context & Emails; blog posts; social media updates; product descriptions \\
\hline
\textbf{Scalenut} & SEO-optimized content & Integrates SEO insights; keyword suggestions; content planning tools & Learning curve for new users; subscription-based; needs manual adjustments & Blog writing; SEO content creation; content strategy development; competitor analysis \\
\hline
\textbf{ChatGPT} & Conversational AI & Maintains context over extended interactions; versatile; high-quality, coherent responses & Produces incorrect or nonsensical answers; lacks real-time data access; potential for biased outputs & Customer service chatbots; virtual assistants; educational tools; content drafting \\
\hline
\textbf{Claude} (2023) \cite{claude} & General writing assistance & Flexible; can handle large inputs; emphasizes safety and reliability in responses & Lacks specialization; availability is limited; less widespread adoption & Creative writing; professional correspondence; content drafting; research assistance \\
\hline
\end{tabular}%
}
\end{table*}

\section{Retrieval-Augmented Generation (RAG)}

RAG is an important advancement in AI text generation that overcomes many limitations of traditional models \cite{lewis2020retrieval}. Unlike conventional language models that rely only on pre-trained data, RAG combines these models with retrieval mechanisms. This allows RAG to access relevant information from external sources like databases or the internet during text generation, improving the accuracy and relevance of the output, especially for up-to-date or specialized information.

Traditional text generation systems create responses based solely on the information stored in the model. In contrast, RAG adds a retrieval step, actively searching for and using external information. This results in more accurate and contextually appropriate responses. By integrating retrieved knowledge, RAG can handle tasks that require real-time information or specific knowledge not included in the model’s training data.

\begin{figure*}
    \centering
    \includegraphics[width=1\linewidth]{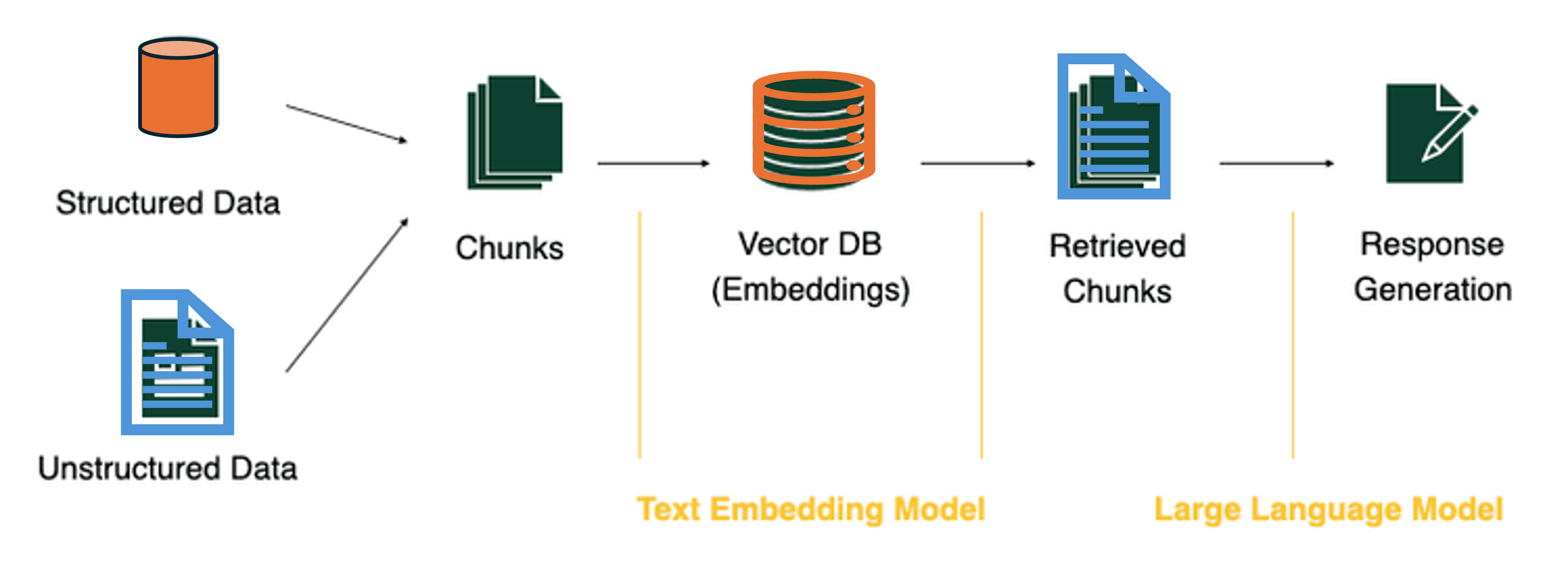}
    \caption{RAG architecture}
    \label{fig:RAG}
\end{figure*}
\subsection{RAG components}
RAG has three main components: (1) retrieval model; (2) embedding model; and (3) generative model. Each component plays a crucial role in producing text that is both accurate and relevant to the user’s query. Table \ref{tab:RAG_components} provides a summary of these.

\subsubsection{Retrieval Model} The retrieval model is responsible for identifying and retrieving relevant documents or data from an external knowledge source. Using techniques like nearest neighbor search, based on pre-encoded document embeddings, the retrieval model efficiently locates pertinent information. For instance, in a customer service scenario, it may retrieve the latest product manual or troubleshooting guide in response to a user query.

\subsubsection{Embedding Model} The embedding model converts both the input query and the retrieved documents into vector representations, or embeddings, to capture their semantic meaning. This step enables the system to match the query with documents that are contextually relevant, beyond simple keyword matching, by understanding the underlying meaning of the text.

\subsubsection{Generative Model} The generative model produces the final output text. After relevant documents are retrieved and embedded, the generative model synthesizes this information to create a coherent, contextually appropriate response. Pre-trained transformers, such as GPT or Text-to-Text Transfer Transformers (T5), are typically used to generate human-like text that integrates both the input query and the retrieved knowledge.

The RAG workflow (Figure \ref{fig:RAG}) involves five stages: (1) chunking divides the dataset into segments; (2) embedding transforms each segment into a vector; (3) vector Database (VectorDB) stores these vectors; (4) retrieval uses VectorDB to find relevant chunks, and (5) response Generation synthesizes these chunks into a coherent answer using an LLM.

\begin{table*}[ht]
\centering
\caption{Components of RAG}
\label{tab:RAG_components}
\begin{tabular}{|l|p{6cm}|p{6cm}|}
\hline
\textbf{Component} & \textbf{Function} & \textbf{Example Use Case} \\
\hline
Retrieval Model & Retrieves relevant documents or data from an external knowledge base. & Customer service: Retrieves product manuals or FAQs. \\
\hline
Embedding Model & Converts input queries and retrieved documents into vector representations for semantic matching. & Question answering: Matches query with related documents. \\
\hline
Generative Model & Generates the final response by combining the input and retrieved knowledge. & Content creation: Generates articles based on user input and external sources. \\
\hline
\end{tabular}
\end{table*}

\section{Tools and Methods for RAG}
RAG combines the power of information retrieval with the ability of generative models to create contextually relevant text. The effectiveness of RAG largely depends on the integration of retrieval mechanisms, the quality of external knowledge sources, and the generative model. Below, we discuss the key components and tools used in RAG.

\subsection{Retrieval Mechanisms}
The retrieval component fetches relevant information or documents that the generative model can use to produce more informed text. Common retrieval methods include:

\begin{itemize}
    \item \textbf{Traditional Search}: Methods such as Term Frequency-Inverse Document Frequency (TF-IDF) and Best Matching 25 (BM25) are used for simpler applications, but they are limited in their ability to understand complex contexts.

    \item \textbf{Embedding-Based Retrieval}: Modern RAG systems rely on embedding-based approaches, where the query and documents are represented as vectors in a high-dimensional space. Techniques like dense retrieval using pre-trained models like BERT and Sentence-BERT are commonly employed.
    \item \textbf{Advanced Search Engines}: Tools like FAISS, Annoy, and Elasticsearch optimize the retrieval process, allowing for efficient search and ranking of large document collections.
\end{itemize}

\subsection{Generative Models}
The generative component is responsible for creating human-like text based on the retrieved information. Key models used in RAG include:

\begin{itemize}
    \item \textbf{GPT-3/4}: These are powerful autoregressive models that generate coherent text by conditioning on retrieved documents.
    \item \textbf{Bidirectional and Auto-Regressive Transformers (BART)}: A transformer-based sequence-to-sequence model that excels in tasks like summarization and question answering, often used in conjunction with retrieval methods.
    \item \textbf{T5}: A flexible model that can be fine-tuned for various text generation tasks, often integrated with retrieval systems for improved generation.
\end{itemize}

\subsection{Knowledge Bases}
The quality of retrieved documents heavily depends on the knowledge sources available. Commonly used knowledge bases include:

\begin{itemize}
    \item \textbf{Wikipedia}: Provides general knowledge and is frequently used in RAG models for various tasks, from question answering to content creation.
    \item \textbf{Domain-Specific Knowledge Bases}: Custom knowledge bases containing specialized information, such as technical manuals, product specifications, or medical data.
    \item \textbf{Real-Time Web APIs}: Services like Google Search API can fetch up-to-date content from the web, providing dynamic knowledge for a generation.
\end{itemize}

\subsection{Evaluation Metrics}
Evaluation of RAG is crucial to ensure both the relevance of the retrieved documents and the quality of the generated text. Metrics like Recall-Oriented Understudy for Gisting Evaluation (ROUGE), Bilingual Evaluation Understudy (BLEU), and F1 Score are used to assess text generation quality, while other domain-specific metrics can be employed for tasks such as factual accuracy.

Table \ref{tab:RAG_tools} gives a summary of these.

\begin{table*}[ht]
\centering
\caption{Tools and Methods for RAG}
\label{tab:RAG_tools}
\begin{tabular}{|l|p{5cm}|p{7cm}|}
\hline
\textbf{Component}  &
\textbf{Methods/Tools} &
\textbf{Description} 
\\
\hline
Retrieval Mechanisms & TF-IDF, BM25, Dense Retrieval, FAISS, Annoy, Elasticsearch & Retrieval methods to fetch relevant documents based on the input query. Dense retrieval uses embedding models like BERT for better accuracy. \\
\hline
Generative Models & GPT-3/4, BART, T5 & Text generation models that create contextually relevant and coherent output conditioned on retrieved documents. \\
\hline
Knowledge Bases & Wikipedia, Custom Knowledge Bases, Real-Time Web APIs (e.g., Google Search API) & External knowledge sources used to provide relevant context or real-time data for generating text. \\
\hline
Evaluation Metrics & BLEU, ROUGE, F1 Score & Metrics for evaluating the quality of the generated text, such as accuracy, fluency, and relevance to the input. \\
\hline
\end{tabular}
\end{table*}

\section{AI Text Detectors (AITD)}

AITD are software tools or algorithms designed to analyze written content to assess the likelihood that it was generated by AI rather than a human \cite{fraser2024detecting}. These use computational techniques to identify patterns, statistical anomalies, or stylistic features typical of AI-generated text. Motivations for AITD include:

\begin{itemize}
    \item \textbf{Academic Integrity}: Preventing plagiarism and ensuring the authenticity of student submissions.
    \item \textbf{Content Moderation}: Detecting automated spam, fake reviews, or disinformation campaigns.
    \item \textbf{Intellectual Property}: Protecting authors and creators from unauthorized use of their work.
    \item \textbf{Security}: Identifying automated phishing attempts or social engineering attacks.
\end{itemize}

This section explores AITD tools up to 2023, providing their evolution, applications, and limitations, with a focus on the latest advancements and emerging challenges in the field.

\begin{itemize}
\item \textbf{GPTZero}: GPTZero, developed by Edward Tian in 2023, detects AI-generated text by analyzing perplexity and burstiness, reflecting variability typical of human writing \cite{gptzero}.
\item \textbf{Turnitin}: Turnitin, a popular plagiarism detector, integrated AI detection in 2023 to analyze linguistic patterns for authenticity, though it sometimes produces false positives \cite{turnitin}.
\item \textbf{ZeroGPT}: ZeroGPT, a free online tool launched in 2023, identifies patterns like repetitive phrasing but has lower accuracy for nuanced texts \cite{zerogpt}.
\item \textbf{GLTR}: The Giant Language Model Test Room (GLTR) uses statistical likelihood to visualize word predictability, helpful in analyzing AI-generated text patterns \cite{gltr}.
\item \textbf{Copyleaks}: Copyleaks, introduced in 2023, uses deep learning to detect AI content across multiple languages but is subscription-based \cite{copyleaks}.
\item \textbf{Crossplag}: Crossplag, launched in 2023, detects AI-assisted plagiarism through machine learning, with limitations on underrepresented languages \cite{crossplag}.
\item \textbf{Hive AI}: Hive AI, launched in 2023, detects AI-generated text and manipulated media, particularly useful for security and media authenticity \cite{hiveai}.
\item \textbf{Scribbr AI Checker}: Scribbr, introduced in 2023, focuses on academic AI plagiarism detection, though with limited language support \cite{scribbr}.
\item \textbf{AI Writing Check}: AI Writing Check is a straightforward, fast tool for detecting AI content in blogs and articles but is less reliable for complex writing \cite{aiwritingcheck}.
\end{itemize}

Table \ref{tab:textgen2} provides a comparison of popular AITD tools.
\begin{table*}[ht]
\caption{Comparison of Popular AITD Tools}
\label{tab:textgen2}
\centering
\renewcommand{\arraystretch}{1.2} 
\setlength{\tabcolsep}{8pt} 
\scriptsize 
\begin{tabular}{|p{2cm}|p{2.5cm}|p{2.5cm}|p{2.5cm}|p{2.5cm}|p{2.5cm}|}
\hline
\textbf{Platform} & \textbf{Learning Type} & \textbf{Primary Focus} & \textbf{Advantages} & \textbf{Limitations} & \textbf{Use Cases} \\
\hline

\textbf{GLTR} \cite{gltr} & Statistical Visualization & Assisting analysis of text predictability & Offers visual insights; educational tool & Requires interpretation; not definitive & Research, education on AI text patterns \\
\hline

\textbf{GPTZero} \cite{gptzero} & Statistical Analysis & Detecting AI-generated essays and assignments & Popular among educators; analyzes text randomness & Inconsistent performance; potential false results & Education sector for academic honesty \\
\hline

\textbf{Turnitin} \cite{turnitin} & Proprietary Algorithms & Integrating AI detection in plagiarism checking & Widely used; seamless integration & Reports of false positives; privacy concerns & Monitoring AI-assisted plagiarism in academia \\
\hline

\textbf{ZeroGPT} \cite{zerogpt} & Pattern Recognition & Providing free AI text detection online & Accessible; easy to use & Less accurate; misclassifies text & General content verification \\
\hline

\textbf{Copyleaks} \cite{copyleaks} & Deep Learning & Multilingual AI text detection & Supports multiple languages; platform integration & Subscription-based; challenges with edited AI text & Education, publishing, business content verification \\
\hline

\textbf{Crossplag} \cite{crossplag} & Machine Learning & Detecting AI-assisted plagiarism & User-friendly reports; multi-language support & Struggles with hybrid texts; language limitations & Academic integrity in educational institutions \\
\hline


\textbf{Hive AI} \cite{hiveai} & Supervised Learning & Detecting AI-generated and manipulated media & Focuses on text and multimedia; robust detection capabilities & Expensive for small-scale users; limited availability of features in free version & Journalism, media authenticity, security analysis \\
\hline

\textbf{Scribbr AI Checker} \cite{scribbr} & Machine Learning & AI plagiarism detection in academic writing & Detailed reports with sources; strong academic focus & Limited language support; misses subtle AI-generated content & Academic writing, thesis and dissertation analysis \\
\hline

\textbf{AI Writing Check} \cite{aiwritingcheck} & Statistical Analysis & Checking text originality and AI authorship & Simple interface; fast detection & Not as reliable for complex or creative AI writing & Content verification for blogs, articles, and news \\
\hline

\end{tabular}
\end{table*}

\section{Ethical Considerations}
\begin{figure}
    \centering
    \includegraphics[width=1\linewidth]{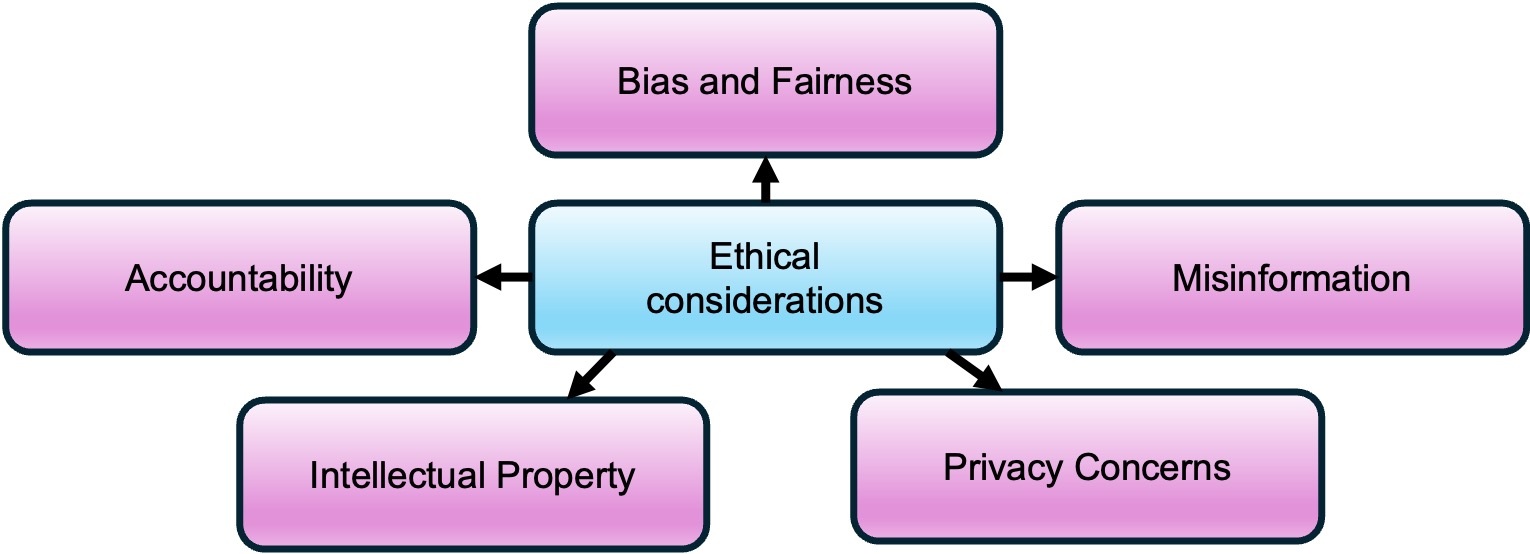}
    \caption{Ethical Considerations}
    \label{fig:ethical}
\end{figure}

This section discusses potential ethical considerations across AITG, RAG, and AITD, as Figure \ref{fig:ethical} highlights these considerations.

\subsection{Bias and Fairness} For AITG, RAG, and AITD models, bias can arise from training data, retrieval sources, or detection criteria, potentially leading to skewed or discriminatory outputs. AITG and RAG models inadvertently reinforce stereotypes in generated responses, while AITDs unfairly flag certain writing styles or demographics. Mitigating these biases requires diversifying datasets, refining retrieval sources, and applying bias detection and fairness audits.

\subsection{Misinformation} AITG and RAG models risk generating or retrieving false information, which could lead to misinformation. Ensuring source reliability, integrating fact-checking, and embedding safeguards in the generation process is key. While AITDs can help detect AI-generated misinformation, they should be complemented with human oversight to reduce false positives and maintain credibility.

\subsection{Privacy Concerns} Privacy is a concern across all three technologies. AITG and RAG models could unintentionally generate sensitive information present in their training or retrieval sources, while AITDs process user data during detection. Complying with data protection standards, anonymizing data where possible, and securing data handling processes are necessary to respect user privacy and legal requirements.

\subsection{Intellectual Property} AITG and RAG models must be cautious about inadvertently replicating copyrighted materials in retrieved or generated text, which raises issues of unlicensed use. AITDs also need to handle copyrighted content sensitively to avoid unauthorized detection. Clear copyright compliance, filtering licensed sources, and monitoring outputs can help address these intellectual property risks.

\subsection{Accountability} Establishing accountability is crucial for tracing and managing errors in these systems. For AITGs and RAGs, transparent tracking of retrieved sources and generation processes is essential to ensure that outputs can be evaluated and errors corrected. For AITDs, protocols are needed for managing false positives, especially when human-authored content is mistakenly labeled as AI-generated, to maintain fairness and accuracy.

\section{Limitations}

\subsection{Lack of Understanding} While AITGs and RAG models produce coherent text, they lack true comprehension. RAG’s retrieval adds context but does not give a deeper understanding, often resulting in responses that appear relevant but can be shallow or occasionally incorrect, especially if the data is outdated. Similarly, AITDs misclassify text as they cannot fully grasp content distinctions, particularly as AI-generated text grows more complex.

\subsection{Hallucinations} Both AITGs and RAG models are prone to hallucinations—producing factually incorrect or invented information. Although RAG models retrieve real-time data, they still reflect biases or inaccuracies from the sources. AITDs face challenges in identifying such hallucinations, struggling to distinguish between AI-generated fabrications and genuine creative work.

\subsection{Dependency on Data Quality} The effectiveness of AITGs, RAG models, and AITDs heavily depends on data quality. RAG’s reliance on external sources means it can amplify inaccuracies from biased or outdated data. AITDs also require comprehensive datasets to avoid missed detections across diverse writing styles, making data quality a critical factor for reliable performance.

\subsection{Resource Intensive} RAG models demand significant computational resources for both retrieval and generation, making them more resource-intensive than standard AITGs. AITDs also require extensive processing to detect AI-generated text accurately. This high resource demand raises concerns around scalability and environmental impact, highlighting the need for optimization in both generative and detection models.

\section{Future work and Conclusion}

AI text generation and detection technologies, including advanced models such as RAG, offer transformative possibilities across sectors such as education, marketing, and customer support. AITGs have enhanced productivity but also raise concerns around bias, privacy, and content authenticity. RAG improves on traditional models by adding contextual relevance through real-time data retrieval, yet it still faces challenges like data quality dependency and potential hallucinations.

Detection tools play a vital role in mitigating these challenges, although they too struggle as AI-generated content grows more sophisticated. RAG’s integration of external knowledge presents both opportunities and new hurdles, emphasizing the need for advances in generative and detection models alike.

Future research should focus on refining both generative and detection technologies, especially RAG, while addressing ethical concerns. Balancing innovation with safeguards will be key to using these tools responsibly and equitably, ensuring that AI remains a force for positive, accountable progress in digital communication, productivity, and creativity.

\bibliographystyle{IEEEtran}
\bibliography{main.bib}

\end{document}